\documentclass{INTERSPEECH2023}
\usepackage{multirow}
\usepackage{float}
\usepackage{tipa}
\usepackage{makecell}
\usepackage{amssymb}
\usepackage{pifont}
\usepackage{adjustbox}
\usepackage[T1]{fontenc}
\newcommand{\cmark}{\ding{51}}
\newcommand{\xmark}{\ding{55}}
\interspeechcameraready

\title{BabySLM: language-acquisition-friendly benchmark\\ of self-supervised spoken language models}

\name{Marvin Lavechin$^{1,2}$, Yaya Sy$^{1}$, Hadrien Titeux$^{1}$, María Andrea Cruz Blandón$^{3}$,\\Okko Räsänen$^{3}$, Hervé Bredin$^{4}$, Emmanuel Dupoux$^{1,2,5}$, Alejandrina Cristia$^{1}$}
\address{
  $^1$LSCP, ENS, EHESS, CNRS, PSL University, Paris, France\quad
  $^2$Meta AI Research, France \\
  $^3$Unit of Computing Sciences, Tampere University, Finland\quad
  $^4$IRIT, CNRS, Toulouse, France\\
  $^5$Cognitive Machine Learning Team, INRIA, France
  \thanks{We thank HPC resources of GENCI-IDRIS (2022-AD011012554); ANR-19-P3IA-0001; J. S. McDonnell Foundation; ERC (ExELang, 101001095). We thank Robin Algayres and Hugo Laurençon for their help designing the syntactic tasks.}
  }
\email{marvinlavechin@gmail.com}

\begin{document}

\maketitle
 
\begin{abstract}
Self-supervised techniques for learning speech representations have been shown to develop linguistic competence from exposure to speech without the need for human labels. In order to fully realize the potential of these approaches and further our understanding of how infants learn language, simulations must closely emulate real-life situations by training on developmentally plausible corpora and benchmarking against appropriate test sets. To this end, we propose a language-acquisition-friendly benchmark to probe spoken language models at the lexical and syntactic levels, both of which are compatible with the vocabulary typical of children's language experiences. This paper introduces the benchmark and summarizes a range of experiments showing its usefulness. In addition, we highlight two exciting challenges that need to be addressed for further progress: bridging the gap between text and speech and between clean speech and in-the-wild speech. 
\end{abstract}

\section{Introduction and related work}

Machine learning for Natural Language Processing (NLP) has led to models that develop linguistic competence from exposure to written or spoken language. On text, Language Models (LMs) now achieve impressive performance on a wide variety of natural language understanding tasks \cite{brown2020language}. More recently, speech-based LMs have also shown impressive linguistic competence on lexical or grammatical acceptability judgment tasks \cite{nguyen2020zero,Dunbar2021TheZR}, or spoken language generation \cite{lakhotia2021generative,Kharitonov2021TextFreePG}. Since these models develop linguistic competence without the need for human labels, they promise to advance our understanding of how infants learn language \cite{dupoux2018cognitive,lavechin2022reverse,warstadt2022artificial}. However, if we want to maximize the impact of the evidence obtained from LMs, it is essential to ensure that our simulations closely emulate real-life situations -- as advocated for syntactic acquisition in text-based LMs in \cite{warstadt2022artificial,huebner2021babyberta}. 

How can we do so? First,  we should match the \textit{quantity} of data available to young infants. Although large differences exist across cultures \cite{cristia2019child} and socioeconomic contexts \cite{dailey2022language}, current estimates of yearly speech input vary between $300$ and $1,000$ hours for American English-learning children \cite{dupoux2018cognitive,hart1995meaningful}. This means that by age 3, American English-learning children would have been exposed to approximately 3,000 hours of speech -- for those who received the most speech input. Yet, by then, infants know many words and already engage in simple conversations \cite{hoff2009blackwell}. Second, we should match the \textit{quality} of data available to young infants. Contrary to LMs, infants do not learn language by scraping the entire web or through exposure to a large quantity of audiobooks. Instead, infants' input is speech -- not text --, and it contains a relatively small vocabulary arranged in simple and short sentences, sometimes overlapping across speakers and laced with various background noises \cite{lavechin2022reverse,lavechinstruggle}.

Evaluating LMs trained on quantitatively and qualitatively plausible corpora requires the creation of adapted benchmarks, but none exists for speech-based LMs -- see \cite{huebner2021babyberta} or the BabyLM challenge \cite{warstadt2023call} for text-based LMs. Current benchmarks using zero-shot probing tasks, although inspired by human psycholinguistics (e.g., spot-the-word or grammatical acceptability judgment tasks), have been designed for models trained on audiobooks \cite{nguyen2020zero}. As a result, these benchmarks use a large vocabulary specific to books (including words like `rhapsodize', `zirconium', or `tercentenary') and probe syntactically complex sentences that are vanishingly rare even in spontaneous adult-adult conversation. 

Here, we propose \textit{BabySLM}, the first language-acquisition-friendly benchmark to probe speech-based LMs at the lexical and syntactic levels, both of which are compatible with the vocabulary typical of children's language experiences. Our benchmark relies on zero-shot behavioral probing of LMs \cite{nguyen2020zero} and considers a spot-the-word task at the lexical level and a grammatical acceptability judgment task at the syntactic level. To show the utility of our benchmark, we first use it to evaluate text-based and speech-based LMs trained on developmentally plausible training sets. The text-based LM is a long short-term memory (LSTM) trained on phonemes or words. The speech-based LM is the low-budget baseline used in the ZeroSpeech 2021 challenge on unsupervised representation learning of spoken language \cite{nguyen2020zero}. Both systems are trained on Providence~\cite{borschinger2013joint}, a dataset of spontaneous parent-child interactions. The comparison between text-based and speech-based LMs shows an important gap that future work should address. Next, \textit{BabySLM} enables us to compare the performance of speech-based LMs when trained on $1,000$ hours of speech extracted from 1) audiobooks, a source of training data commonly used \cite{kahn2020libri,mohamed2022self}; or 2) child-centered long-form recordings acquired via child-worn microphones as people go about their everyday activities \cite{bergelson2019north}. Our results reveal that speech-based LMs are overly sensitive to the differences between clean speech and in-the-wild speech.

\section{Methods}

\subsection{Metrics}


\subsubsection{Lexical evaluation: the spot-the-word task}

\begin{table}[tbh]
  \caption{\textbf{Lexical task.} Minimal pairs of real and pseudo-words. Phonetic (Phon.) transcriptions are given in International Phonetic Alphabet (IPA) standard. Orthographic (Orth.) transcriptions of pseudo-words are proposed for ease of reading.}
  \label{tab:lexical}
  \centering
  \begin{adjustbox}{width=\columnwidth}
  \begin{tabular}{c l l c l l}
    \toprule
    & \multicolumn{2}{l}{\textbf{Pseudo-words}} &  & \multicolumn{2}{l}{\textbf{Pseudo-words}}\\
    \textbf{Word} & \textbf{Phon.} & \textbf{Orth.} & \textbf{Word} & \textbf{Phon.} & \textbf{Orth.}\\
    \midrule
    \multirow{5}*{\makecell{hello\\ \textipa{h~@~l~oU}}} & \textipa{l~@~l~oU} & lello & \multirow{5}*{\makecell{thanks \\ \textipa{T~\ae~N~k~s}}} & \textipa{T E N k s} & thaynks\\
    & \textipa{p~@~l~oU} & pello & & \textipa{T~O~N~k~s} & thoanks\\
    & \textipa{s~@~\*r~oU} & sero & & \textipa{T~I~s~k~s} & thisks\\
    & \textipa{d~@~l~oU} & dello & & \textipa{T~\ae~m~p~s} & thamps\\
    & \textipa{s~@~l~oU} & sello & & \textipa{T~\ae~n~t~s} & thants\\
    \midrule
    \multirow{5}*{\makecell{cookie\\ \textipa{k~U~k~i:}}} & \textipa{k~U~t~i:} & kootie & \multirow{5}*{\makecell{jump \\ \textipa{dZ~2~m~p}}} & \textipa{dZ~\ae~m~p} & jamp\\
    & \textipa{k~U~n~i} & koonie & & \textipa{dZ~2~l~k} & julk\\
    & \textipa{\*r~U~d~i:} & roodie & & \textipa{dZ~2~s~k} & jusk\\
    & \textipa{\*r~U~t~i:} & rootie & & \textipa{dZ~2~f~t} & juft\\
    & \textipa{\*b~U~n~i:} & boonie & & \textipa{dZ~2~b~s} & jubs\\
    \bottomrule
  \end{tabular}
  \end{adjustbox}
\end{table}

\noindent \textbf{General principle.} In the lexical task, the system is presented with minimal pairs of an existing word and a pseudo-word that is phonologically plausible but does not actually exist \cite{nguyen2020zero,le2017comparing} (examples in Table \ref{tab:lexical}). The system gets a score of $1$ if it returns a higher probability for the former, and $0$ otherwise. Contrary to~\cite{nguyen2020zero}, we generate multiple pseudo-words per word. Scores are first averaged across pseudo-words to yield per-word accuracy, which are then averaged across all words to yield a measure of \textit{lexical accuracy}.

\noindent \textbf{Task generation.} We first listed all words in the American English CHIld Language Data Exchange System (CHILDES) database \cite{macwhinney1985child}. This database contains human-annotated transcripts of various child-centered situations (play sessions, storytelling, etc.), making it a valuable source of vocabulary in real children's input. After excluding items not found in either the Celex \cite{baayen1996celex2} or CMU dictionary \cite{weide1998carnegie} (e.g., mispronounced, incorrectly annotated or made-up words: `insectasaurus', `hiphippopotamus'), we obtained $28,000$ word types. Pseudo-words were produced using the Wuggy pipeline \cite{keuleers2010wuggy}, which generates, for a given word, a list of candidate pseudo-words matched for syllabic and phonotactic structure. We applied the same post-processing steps used in \cite{nguyen2020zero}. Contrary to \cite{nguyen2020zero}, to ensure that there is no bias from phone-based unigrams or bigrams, we balanced the count of pseudo-words that had higher (or lower) phonemes unigram and bigram probabilities compared to those computed for the actual word. If a given word had only pseudo-words with higher (or lower) unigram or bigram possibilities, it was discarded from the evaluation set. The resulting $>90,000$ minimal pairs across $18,000$ words were each synthesized using Google Text-To-Speech (TTS) system using $10$ voices ($5$ males, $5$ females). 

\subsubsection{Syntactic evaluation: grammatical acceptability}

\begin{table}[tbh]
  \caption{\textbf{Syntactic task.} Minimal pairs of grammatical (\cmark) and ungrammatical (\xmark) sentences from each of the six syntactic phenomena included in our benchmark. N is the number of $1,000$ minimal pairs within each category.}
  \label{tab:syntactic}
  \centering
  \begin{tabular}{c c l}
    \toprule
    \textbf{Phenomenon} & \textbf{N} & \textbf{Sentence example}\\
    \midrule
    \multirow{2}*{\makecell{Adjective-noun\\ order}} & \multirow{2}*{$1.6$} & \cmark~~\textit{The good mom.}\\
    & & \xmark~~\textit{The mom good.}\\
    \midrule
    \multirow{2}*{\makecell{Noun-verb\\ order}} & \multirow{2}*{$1$} & \cmark~~\textit{The dragon says.}\\
    & & \xmark~~\textit{The says dragon.}\\
    \midrule
    \multirow{2}*{\makecell{Anaphor-gender\\ agreement}} & \multirow{2}*{$2$} & \cmark~~\textit{The dad cuts himself.}\\
    & & \xmark~~\textit{The dad cuts herself.}\\
    \midrule
    \multirow{2}*{\makecell{Anaphor-number\\ agreement}} & \multirow{2}*{$1$} & \cmark~~\textit{The boys told themselves.}\\
    & & \xmark~~\textit{The boys told himself.}\\
    \midrule
    \multirow{2}*{\makecell{Determiner-noun\\ agreement}} & \multirow{2}*{$3.6$} & \cmark~~\textit{Each good sister.}\\
    & & \xmark~~\textit{Many good sister.}\\
    \midrule
    \multirow{2}*{\makecell{Noun-verb\\ agreement}} & \multirow{2}*{$1.6$} & \cmark~~\textit{The prince needs the princess.}\\
    & & \xmark~~\textit{The  prince need the princess.}\\
    \bottomrule
  \end{tabular}
\end{table}

\noindent \textbf{General principle.} In the syntactic task, the system is presented with minimal pairs of grammatical and ungrammatical sentences across six syntactic phenomena \cite{nguyen2020zero,huebner2021babyberta} (examples in Table \ref{tab:syntactic}), giving the system a score of $1$ when it assigns a higher probability to the former, and $0$ otherwise. We average scores within each syntactic phenomenon, then across phenomena to obtain our measure of \textit{syntactic accuracy}. 

\noindent \textbf{Task generation.} We generated templates for each of the six syntactic phenomena explored. For instance, for the noun-verb agreement phenomenon, we used templates such as ``The <noun$_1$> <3$^{rd}$ person verb> <noun$_2$>" versus ``The <noun$_1$> <1$^{st}$ person verb> <noun$_2$>". Contrary to \cite{nguyen2020zero}, we restricted this benchmark to simple syntactic phenomena and short sentences which better reflect the type of input children are exposed to. We filled the templates using high-frequency words from CHILDES \cite{macwhinney1985child}. For instance, selected animate nouns include words like `mom', `girl', or `cat';  selected adjectives include words like `good', `little', or `big'; and selected verbs include words like `see', `know', or `need'. The resulting $10,800$ minimal pairs were each synthesized using Google TTS system using the same $10$ voices ($5$ males, $5$ females).

\subsubsection{Development and test split}

For both our lexical and syntactic evaluation sets, we randomly selected one male and one female voice for the development set and the $8$ remaining ones for the test. We randomly selected \SI{20}{\percent} of the lexical and syntactic minimal pairs for the development set and the remaining \SI{80}{\percent} for the test.

\subsection{Training sets}

We built a first training set by extracting human-annotated speech utterances from Providence \cite{borschinger2013joint}, a publicly available corpus containing transcribed recordings of six American children during spontaneous interactions with their parents. Available utterance-level timestamps were refined with a pretrained voice activity detection (VAD) system \cite{Lavechin2020AnOV}. We converted human orthographic transcripts into phonetic transcripts using~\cite{bernard2021phonemizer}. This procedure resulted in $128$ hours of highly naturalistic infant-parent interactions in audio, orthographic, and phonetic form, allowing us to compare LMs trained on speech, phonemes, or words. 

We built a second training set by extracting $1,024$ hours of adult speech utterances -- using the same VAD system \cite{Lavechin2020AnOV} -- from SEEDLingS \cite{bergelson2019north}, a corpus of child-centered long-form recordings collected in $61$ American English families. This training set enables us to train speech-based LMs in maximally plausible conditions, i.e., directly on what infants hear.

\subsection{Models}

\begin{table*}
  \caption{\textbf{The BabySLM benchmark.} Lexical and syntactic accuracies obtained by different language models trained on developmentally plausible corpora of speech, phonemes, or words. Numbers are computed on the test set, and performances on the development set are reported using small font size. The starred cumulated duration and number of words are estimates based on the \SI{1.2}{M} of words present in the $128$ hours of speech from Providence. Data plausibility indicates the extent to which the training set is close to the real sensory signal available to infants.}
  \label{tab:results}
  \centering
  \begin{tabular}{l c c c c c c c}
    \toprule
    & & & \textbf{Cumulated} & \textbf{Number of} & \textbf{Data} & \textbf{Lexical} & \textbf{Syntactic}\\
    
    \textbf{System} & \textbf{Input} & \textbf{Training set} & \textbf{duration ({\boldmath \unit{\hour}})} & \textbf{words ({\boldmath \unit{M}})} & \textbf{plausibility} & \textbf{acc. ({\boldmath \unit{\percent}})} & \textbf{acc. ({\boldmath \unit{\percent}})}\\
    \midrule
    Random baseline & --- & --- & $0$ & $0$ & --- & $49.2$ \scriptsize{$52.5$} & $49.3$ \scriptsize{$50.0$}\\
    STELA \cite{lavechin2022statistical} & speech & SEEDLingS & $1024$ & ~~$9.6^\star$  & +++ & 49.5 \scriptsize{$45.4$} & $50.3$ \scriptsize{$50.5$}\\
    STELA \cite{lavechin2022statistical} & speech & Providence & $128$ & $1.2$  & ++ & $56.8$ \scriptsize{$47.1$} & $50.3$ \scriptsize{$51.1$}\\
    LSTM  & phonemes & Providence & $128$ & $1.2$ & + &$75.4$ \scriptsize{$75.2$} & $55.1$ \scriptsize{$55.9$}\\
    LSTM &  words (BPE) & Providence & $128$ & $1.2$  & + & ---  & $65.1$ \scriptsize{$65.3$} \\
    BabyBERTa \cite{huebner2021babyberta} & words (BPE) & AO-CHILDES  & ~~$533^\star$ & $5$  & + & --- & $70.4$ \scriptsize{$70.4$}\\
    \bottomrule
  \end{tabular}
\end{table*}

\noindent \textbf{STELA (speech-based).} STELA is a speech-based LM originally proposed in \cite{nguyen2020zero,lavechin2022statistical}. It comprises an acoustic model that learns discrete representations of the audio and a language model trained on top of the learned discrete representations. The acoustic model is built from a Contrastive Predictive Coding (CPC) model followed by a K-means clustering algorithm. The language model consists of LSTM layers. We used the same architecture and hyper-parameters as the low-budget baseline proposed in \cite{nguyen2020zero}. Contrary to \cite{nguyen2020zero} who trained CPC by sampling the positive and negative examples from the same speaker, we applied a second constraint: negative examples were drawn from temporally close speech sequences to reduce mismatch between the positive and negative examples in terms of their local environment as this was found to be helpful when training on long-forms~\cite{lavechinstruggle}.

\noindent \textbf{LSTM (text-based).} We include LSTM LMs trained on words -- using byte-pair encoding -- or on phonemes, using the same architecture and hyper-parameters than \cite{nguyen2020zero}.

\noindent \textbf{BabyBERTa (text-based).} BabyBERTa \cite{huebner2021babyberta} is a transformer-based LM trained on a \SI{5}{M} word corpus of American English child-directed input built from the CHILDES database \cite{macwhinney1985child}.

\section{Results and discussion}
\label{sec:res}

\subsection{The BabySLM benchmark}

Results obtained on our \textit{BabySLM} benchmark are reported in Table \ref{tab:results}. Rows are sorted according to the plausibility of the training data. Child-centered long-form recordings (SEEDLingS) have the highest plausibility score as these recordings faithfully capture children's everyday language experiences. In particular, long-forms collect audio data over a whole day -- or several -- and therefore sample the full range of language experiences across all possible contexts: the child may be in or out of the house, the speech may be directed to the child or others, etc. The audio extracted from in-home recordings of spontaneous infant-parent interactions (Providence) is slightly less plausible as it fails to capture the full range of language experiences: fewer speakers than in a real-life setting, most of the speech is directed to the child, etc. Finally, words and phonemes extracted from AO-CHILDES or Providence have the lowest plausibility score since infants do not learn language from orthographic or phonetic transcriptions but from the continuous signal that is speech.

Results indicate no evidence of lexical and syntactic knowledge for STELA trained on $1,024$ hours of speech from SEEDLingS. This contrasts, in appearance, with what has been found in the ZeroSpeech challenge \cite{nguyen2020zero}, but this is due to the large variability of speech found in long-forms as we will see in Section \ref{sec:inthewild}. Results are no different for STELA trained on $128$ hours of speech extracted from Providence whose lexical and syntactic accuracies remain close to chance level. However, we hypothesize that the lexical accuracy obtained by STELA might increase with more audio data from semi-controlled recordings of infant-parent interactions as these contain cleaner speech than what is typically found in long-forms. Contrary to speech-based LMs, text-based LMs perform largely above chance level. As expected, the LSTM model trained on words reaches higher syntactic accuracy than the LSTM trained on phonemes. The highest syntactic accuracy is obtained by BabyBERTa, which is a transformer-based LM and has been trained on a larger quantity of data than our LSTM LMs.

Performances on \textit{BabySLM} show a clear gap between text-based and speech-based LMs. Another important finding is that, as of now, spoken language modeling from children's real language experiences seems out of reach, as evidenced by the chance-level lexical and syntactic accuracies obtained by STELA trained on SEEDLingS. We dedicate the remaining sections to illustrating these two challenges: bridging the gap between text and speech and between clean speech and in-the-wild speech.

\subsection{Language modeling: from text to speech}

\begin{figure}
\centering
\includegraphics[width=.8\columnwidth]{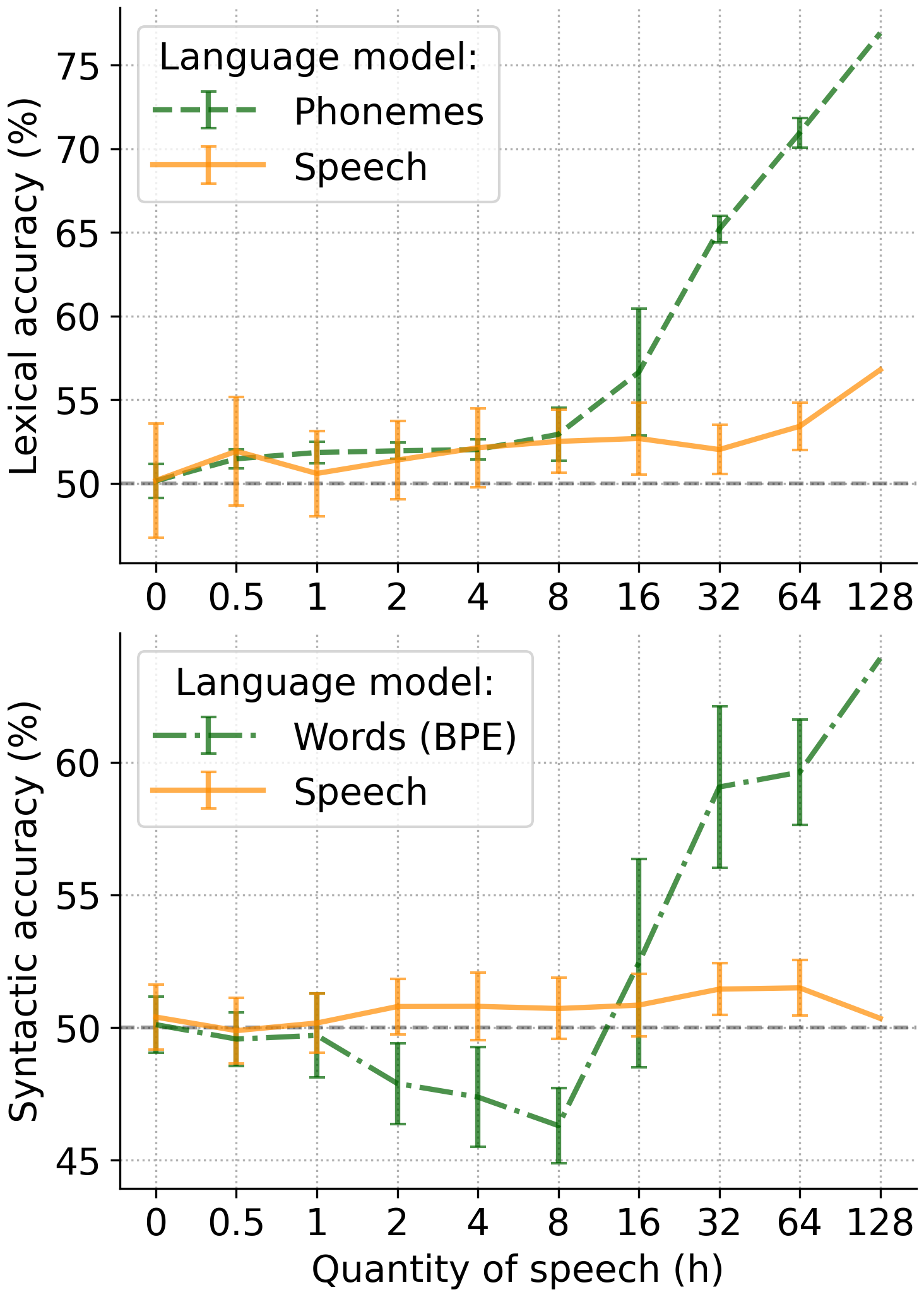}
\caption{\textbf{Language modeling from text to speech.} Top panel shows the lexical accuracy obtained by language models trained on audio (STELA) or phonemes (LSTM). Bottom panel shows the syntactic accuracy obtained by language models trained on audio (STELA) or byte-pair-encoded (BPE) words (LSTM). All models are trained on the Providence corpora in audio, phonetic, or orthographic form. Numbers are computed on the test set. Error bars represent standard errors computed across mutually exclusive training sets.}
\label{fig:text_to_speech}
\vspace{-.3cm}
\end{figure}

\noindent Figure \ref{fig:text_to_speech} shows lexical and syntactic accuracies obtained by text-based (words or phonemes) or speech-based LMs as a function of quantity of data. The LSTM trained on phones requires at least $16$ hours of speech, equivalent to $150,000$ words, to start performing above chance level. Once lexical knowledge has emerged, the model follows a logarithmic trend (note the log-scale x-axis), initially improving rapidly and then slowing down. In other words, we need to double the amount of data to obtain the same gain in lexical accuracy. The same patterns hold for the syntactic accuracy obtained by the LSTM model trained on words\footnote{Note, however, that the syntactic accuracy obtained by the LSTM model trained on words decreases to \SI{45}{\percent} (below chance level) between $0$ and $8$ hours ($=75,000$ words). This effect was found to be driven by co-occurrence statistics in the noun-verb order task. The same pattern was found with a $3$-gram model, with a slight decrease between $0$ and $8$ hours and an increase between $8$ and $128$ hours.}. For STELA, the lexical accuracy remains close to chance level, although the curve seems to increase between $32$ and $128$ hours of speech, and there is no evidence for syntactic knowledge.

All in all, the lexical and syntactic accuracy slopes show very different patterns when training from raw speech or phonemes or words. This is despite receiving the same data in different forms. Admittedly, the speech-based LM faces a more challenging task as it must learn its own discrete units, while text-based LMs must not. Future work might investigate how these slopes change with more data, particularly for the speech-based LM for which $128$ hours seems insufficient.

\subsection{Language modeling: from clean to in-the-wild speech}
\label{sec:inthewild}

\begin{figure}
\centering
\includegraphics[width=.75\columnwidth]{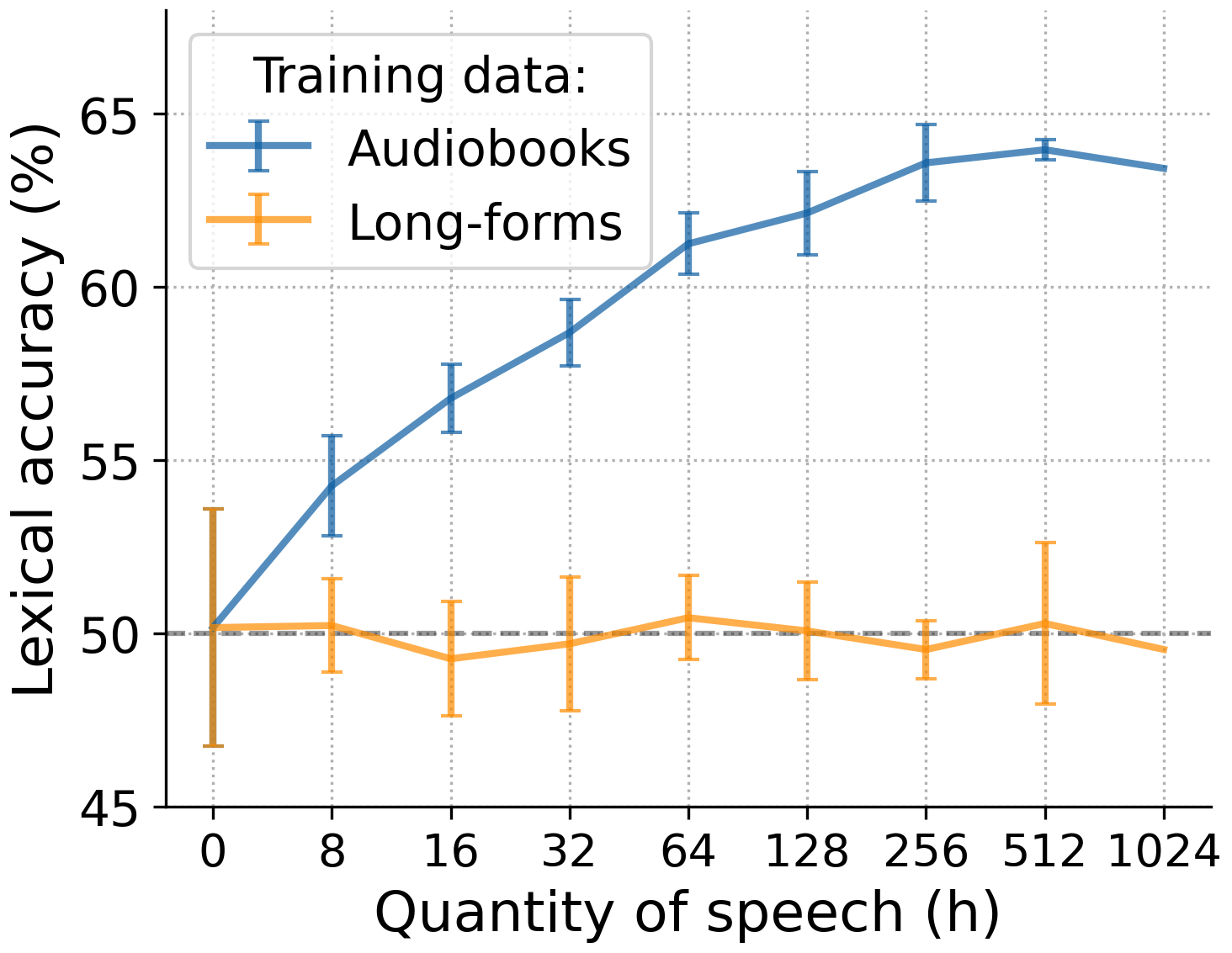}
\caption{\textbf{Language modeling from clean to in-the-wild speech.} Lexical accuracy obtained by STELA trained on audiobooks (Libri-light, in blue) or child-centered long-forms (SEEDLingS, in orange) as a function of speech quantity. Numbers are computed on the test set. Error bars represent standard errors computed across mutually exclusive training sets.}
\label{fig:cleanwild}
\vspace{-.3cm}
\end{figure}

So far in the paper, we have little evidence that lexical or syntactic knowledge can emerge in speech-based LMs. To address this concern, we ran one more experiment, this time training STELA under more controlled recording conditions: on up to $1,024$ hours of speech extracted from audiobooks -- commonly used to train speech-based LMs \cite{kahn2020libri}. Figure \ref{fig:cleanwild} compares this experiment against the performance obtained by STELA when trained on child-centered long-forms (SEEDLingS, Table \ref{tab:results}). 

Results are unequivocal: we observe a strong improvement on the lexical task for the model trained on audiobooks, while the same model trained on long-forms remains at chance level. On the syntactic task (not shown above), STELA trained on $1,024$ hours of audiobooks obtains an accuracy of \SI{52.8}{\percent} compared to \SI{50.3}{\percent} on long-forms. This is in line with the results in~\cite{nguyen2020zero} showing that more powerful architectures are necessary to learn at the syntactic level.

Why do we observe chance-level performance when training on long-forms? First, the speech signal found in long-forms is much more challenging than the one found in audiobooks: the speech might be distorted as it is being spoken far from the child; it might overlap with various background noises; and it is often produced in short turns that might be under-articulated -- see \cite{lavechinstruggle} for a comparative analysis. Another essential factor to consider is the domain mismatch between the training and test sets. While the training set contains far-field under-articulated speech as well as close-field storytelling, the test set consists of well-articulated synthesized stimulus to which STELA fails to generalize. However, infants show no difficulties generalizing from uncontrolled real-life conditions to more controlled ones (in-laboratory conditions). We advocate here that generalization is part of the language acquisition problem, and LMs should be evaluated accordingly. 

We hypothesize that the discrete units learned by STELA might be too dependent on the various non-linguistic factors found in long-forms, as suggested in \cite{lavechinstruggle}. This dependency could prevent the LSTM LM from learning long-term dependencies necessary to solve the lexical or syntactic tasks.

\section{Conclusion}

Benchmarks are instrumental in allowing cumulative science across research teams. In this paper, we have described how BabySLM has been carefully designed to be adapted to the kinds of words and sentences children hear. We have shown how it can be used to evaluate LMs trained on developmentally plausible text or speech corpus. By doing so, we revealed two outstanding challenges that the community must solve to build more plausible cognitive models of language acquisition. First, we need to reduce the gap between text-based and speech-based LMs, as the latter performed close to chance level on BabySLM. Second, we need to reduce the gap between LMs trained on clean and in-the-wild speech, as evidenced by the striking difference we obtained on the lexical task when training on clean audiobooks versus ecological long-forms. 

Future work might consist in evaluating speech-based LMs grounded in the visual modality \cite{nikolaus2022learning}, or linking performances obtained on \textit{BabySLM} with behavioral measures in infants -- e.g., age of acquisition as in \cite{portelance2020predicting}. A crucial limitation of our benchmark is that it focuses on English, which already accounts for a whopping \SI{54}{\percent} of language acquisition studies \cite{kidd2022diverse}. We hope that this paper, together with shared scripts\footnote{\texttt{https://github.com/MarvinLvn/BabySLM}}, will facilitate the creation of similar benchmarks in other languages.

\newpage
\bibliographystyle{IEEEtran}
\bibliography{mybib}

\end{document}